\documentclass[default, iicol, sn-mathphys, Numbered]{sn-jnl}

\usepackage{graphicx}%
\usepackage{multirow}%
\usepackage{amsmath,amssymb,amsfonts}%
\usepackage{amsthm}%
\usepackage{mathrsfs}%
\usepackage[title]{appendix}%
\usepackage{xcolor}%
\usepackage{caption}
\usepackage{array}
\usepackage{amsmath}
\usepackage{mathtools}
\usepackage{amssymb}
\usepackage{arydshln}
\usepackage{textcomp}%
\usepackage{manyfoot}%
\usepackage{booktabs}%
\usepackage{algorithm}%
\usepackage{algorithmicx}%
\usepackage{subcaption}
\usepackage{algpseudocode}%
\usepackage{graphicx}
\usepackage{listings}%

\theoremstyle{thmstyleone}%
\theoremstyle{thmstyletwo}%

\theoremstyle{thmstylethree}%

\newcommand{\mb}[1]{\mathbf{#1}}
\newcommand{\bs}[1]{\boldsymbol{#1}}

\begin{document}

\title[Article Title]{Teleoperation in Robot-assisted MIS with Adaptive RCM via Admittance Control}


\author[1]{\fnm{Ehsan} \sur{Nasiri}}\email{enasiri@stevens.edu}
\author[1]{\fnm{Srikarran} \sur{Sowrirajan}}\email{ssowrira@stevens.edu}
\author*[1]{\fnm{Long} \sur{Wang}}\email{lwang4@stevens.edu}


\affil[1]{\orgdiv{Department of Mechanical Engineering}, \orgname{Stevens Institute of Technology}, \orgaddress{\street{1 Castle Point Terrace}, \city{Hoboken}, \state{NJ}, \postcode{07030}, \country{US}}}


\abstract{This paper presents the development and assessment of a teleoperation framework for robot-assisted minimally invasive surgery (MIS). The framework leverages our novel integration of an adaptive remote center of motion (RCM) using admittance control. This framework operates within a redundancy resolution method specifically designed for the RCM constraint. We introduce a compact, low-cost, and modular custom-designed instrument module (IM) that ensures integration with the manipulator, featuring a force/torque sensor, a surgical instrument, and an actuation unit for driving the surgical instrument. The paper details the complete teleoperation framework, including the telemanipulation trajectory mapping, kinematic modelling, control strategy, and the integrated admittance controller. Finally, the system's capability to perform various surgical tasks was demonstrated, including passing a thread through the rings, picking and placing objects, and trajectory tracking.}





\maketitle



\section{Introduction}\label{Introduction}

\begin{figure}
	\centering
	\includegraphics[width=0.5\textwidth]{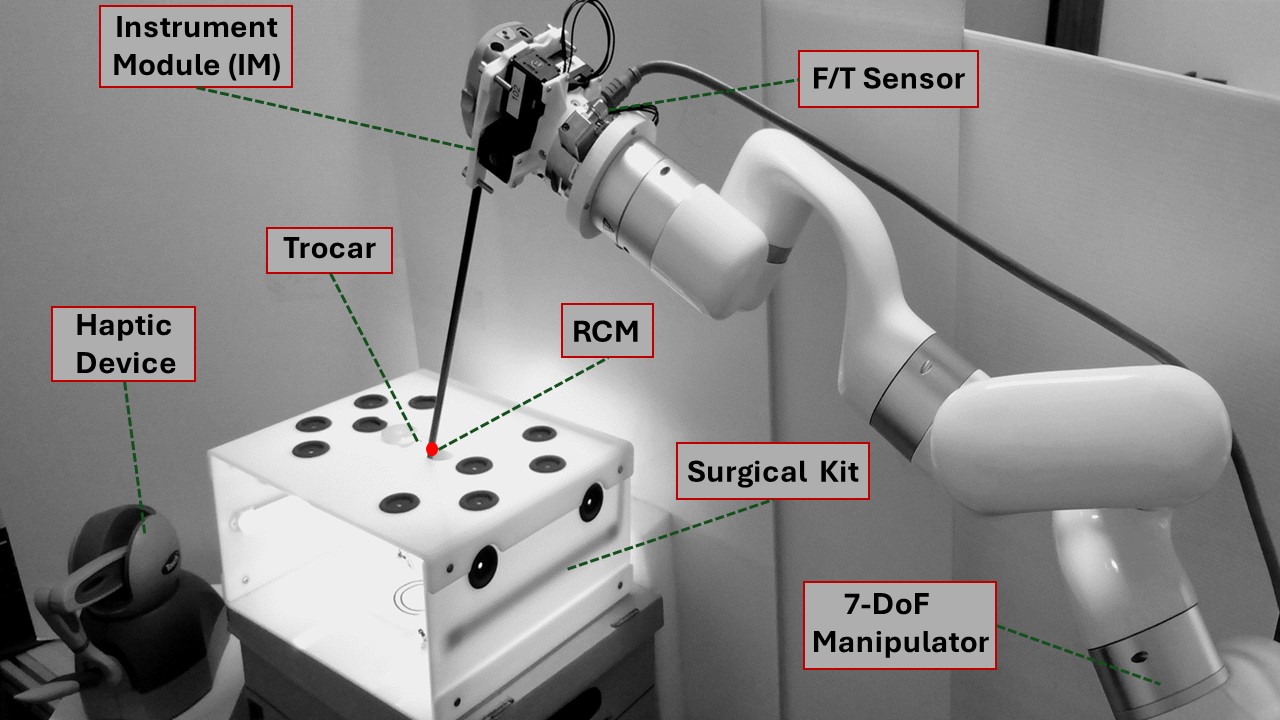}
	\caption{System architecture includes a 7-DoF manipulator, custom-designed Instrument Module (IM), haptic stylus device, and the surgical kit}
	\label{fig:System_arch}
\end{figure}
Minimally invasive surgery (MIS) offers patients a less invasive alternative to traditional open surgery. A few advantages of MIS is reduced postoperative pain, shorter hospital stays, and faster recovery times \cite{mis_review1}. MIS has seen widespread adoption across various surgical specialties, including urology, gynecology, and general surgery, among others \cite{mis-urology, mis-gynecology, mis-cardiac}. This paradigm shift in surgical practice has been fueled by advancements in technology, such as robot-assisted MIS, which provide surgeons with enhanced visualization, precision, and dexterity. The past few decades have seen significant research and collaboration towards robot-assisted surgical technology \cite{kwoh1988, Sackier, Ho}. In \cite{mis-lap-vs-robot}, robotic surgery demonstrated a better overall surgical performance score and shorter operative time than laparoscopy during minimally invasive pancreatoduodenectomy. Furthermore, robot-assisted surgeries have been successfully implemented to laparoscopic colon and rectum resection procedures \cite{Antoniou2012, Hubens2003}. 

Robot-assisted MIS is carried out by making small incisions on the patients body to allow the insertion of surgical tools and cameras that are guided and manipulated from outside the body. The insertion points for the tools are created using a surgical tool called the trocar, which punctures the body and acts as a port of access for the surgical tools. A primary objective of robot-assisted surgery is manipulation at the tip of the surgical instrument, which is subject to motion constraints at the trocar called the remote center of motion (RCM). The RCM constraint only allows motion of the surgical tools to translate along its axis and rotate about the trocar. Robotic systems incorporating the RCM constraint in their control algorithms require additional degrees of freedom (DoF) to maintain the same dexterity as traditional laparoscopic surgery \cite{Azimian, Sadeghian}.
The RCM constraint is programmed and a task priority method is developed to address the system kinematic control in \cite{Aghakhani}. Meanwhile, instrument positioning and RCM control was assessed using a dual quaternion-based kinematic controller in \cite{Marinho2014APR}. In \cite{Locke2007}, an isotropic-based kinematic optimization is used to select the location of the RCM. A vision-based calibration method leveraging perspective projection geometry of the endoscopic camera on dual RCM-based robots during human-robot collaborative MIS was presented in \cite{wangzerui2018}.

Previous works such as \cite{Azimian} and \cite{Aghakhani} assume the RCM as a fixed point. However, real-world conditions have significant challenges in maintaining a fixed RCM location. Conditions such as the patient's heartbeat and breathing create continuous variations in the trocar location, with displacements of a few centimeters occurring in the chest and abdominal areas \cite{Lowanichkiattikul2016, Riviere}. 
To account for a moving trocar in the control of the robotic system, visual sensors can be used to estimate the trocar velocity, as discussed in \cite{Sadeghian}. This approach assumes the availability of additional sensors to provide this velocity information.
Force feedback has also been used to determine the changing position of the trocar as described in \cite{fontanelli2020external}, where the force sensors are placed at the trocar. While this solution offers direct sensing of the interaction forces at the trocar, it poses some limitations with regards to sterilization and capital equipment maintenance. In our previous work, we incorporated force/torque sensing at the base of the instrument module (IM) to derive the interaction forces at the RCM~\cite{nasiri2024}.

In robot-assisted MIS, the patient side manipulator (PSM) system is teleoperated by a surgeon from a haptic console. The console enables the surgeon to directly control the PSM system in a responsive manner, translating their hand movements into precise movements of the surgical instruments inside the anatomy. The console can also provide feedback to the surgeon regarding the forces being applied during the procedure, which could aid the surgeon to adjust their movements in real-time. An inexpensive open-source haptic stylus, \emph{3DSystems Touch} \cite{3dstouch}, has been used to deploy teleoperation of surgical robotic systems to perform transnasal surgery in \cite{burgner2014}. A leader-follower trajectory mapping for the teleoperation of a single port access surgical robot using the 3DSystems Touch as the leader was presented in \cite{bajo-IREP}. Due to its widespread adoption \cite{Chen2020, munawar2022, Long2023}, we integrate the 3D Systems Touch device in this work to enable teleoperation of a Patient-Side Manipulator (PSM) system. This system includes a 7-DoF robot manipulator equipped with a 4-DoF da Vinci Research Kit (dVRK) wristed surgical instrument (similar to the one used in \cite{fontanelli2017modelling, fontanelli2018v}), as shown in Fig.~\ref{fig:System_arch}.

The contributions of this research paper are as follows:
\begin{itemize}
    \item Development of a teleoperation framework for robot-assisted MIS. The framework leverages an adaptive admittance control accommodating the moving RCM constraint.
    \item Development of an open-source Instrument Module (IM) and its integration with a 7-DoF manipulator. The IM includes an F/T sensor, a 4-DoF dVRK wristed surgical instrument, and an actuation unit to drive the surgical instrument.
    \item Validation of the proposed system through simulations and hardware experiments. Validated tasks include trajectory tracking, pick-and-place, and thread passing.
\end{itemize}

The outline of this paper is as follows. In Section~\ref{Kinematic}, we explain the kinematic models for the RCM constraint and the integration of an admittance controller into the redundancy resolution method to handle the RCM constraint. This is followed by a discussion of the teleoperation trajectory mapping and resolved motion-rate controller in Section~\ref{haptic}, along with the system control architecture. Section~\ref{IDM} describes the mechatronics design of the open-source customized IM. Section~\ref{validation} describes both the simulation and hardware experimental validations of the proposed framework. Finally, in Section~\ref{conclusion}, we discuss the results and future work.

\begin{table}[ht]
	\centering
	\caption{Nomenclature for Kinematics}
	\label{tab:nomenclature}
	{\renewcommand{\arraystretch}{1.3}
		\footnotesize
		\begin{tabular}{m{.15\columnwidth} m{.65\columnwidth}}
			\hline
			Symbol & \vspace{1mm} Description \vspace{1mm} \\
			\hline
			$\mb{{p}}_{\text{rcm}}$
			& \vspace{1mm} RCM position \\
			\hdashline
			$\mb{{p}}_{\text{end}}$
			&\vspace{1mm} Robot \emph{end-effector} position\\
			\hdashline
			$\mb{{p}}_{\text{ins}}$
			&\vspace{1mm} Wrist center position of the surgical \emph{instrument}  \\
			\hdashline
			$\mb{d}_\text{ins}$
			&\vspace{1mm}  Vector that points from $\mb{{p}}_{\text{end}}$ to $\mb{{p}}_{\text{ins}}$ \\
			\hdashline
			$\mb{{J}}_{\text{rcm}}$
			&\vspace{1mm}  Jacobian for the RCM  \\
			\hdashline
			$\mb{{J}}_{\text{end}}$
			&\vspace{1mm}  Jacobian for the end-effector  \\
			\hdashline
			$\mb{{J}}_{\text{ins}}$
			&\vspace{1mm} Instrument Jacobian   \\
                \hdashline
			$\mb{{q}}_{\text{aug}}$
			&\vspace{1mm} Augmented joint space vector   \\
			\hdashline
			$\mb{K}_{\text{adm}}$, $\mb{K}_{\text{adm}\perp}$
			&\vspace{1mm}   A scalar admittance gain and it's projection \\  
			\hdashline
			$\mb{{f}}_{\text{rcm}}$
			&\vspace{1mm}  Actual force exerted from patient's body at RCM point \\
			\hdashline
			$\hat{\mb{{f}}}_{\text{rcm}}$
			&\vspace{1mm}  Estimated Force exerted from patient's body at RCM point by F/T sensor  \\
			\hdashline
			$\mb{{f}}_{\text{e,rcm}}$
			&\vspace{1mm}  RCM force tracking error\\
			\hdashline
			$\bs{\Omega}$
			&\vspace{1mm}   The constructed projection matrix  \\
			\hdashline
			${\bs{{\xi}}}_{\text{aug}}$
			&\vspace{1mm}  Augmented command velocity vector \\
			\hdashline
                ${\mb{{x}}}_{\text{ins}}$
			&\vspace{1mm}  Pose of the robot \textit{instrument} \\
			\hdashline
			$\lambda$
			&\vspace{1mm}   The interpolation variable of RCM point along the instrument shaft, $\lambda\in(0,1)$.  \\
			\hline
		\end{tabular}
	}
 \label{tab:nomeclauture}
\end{table}
\begin{figure}[!th]
    \centering
    \includegraphics[width=0.48\textwidth, height=0.36\textheight]{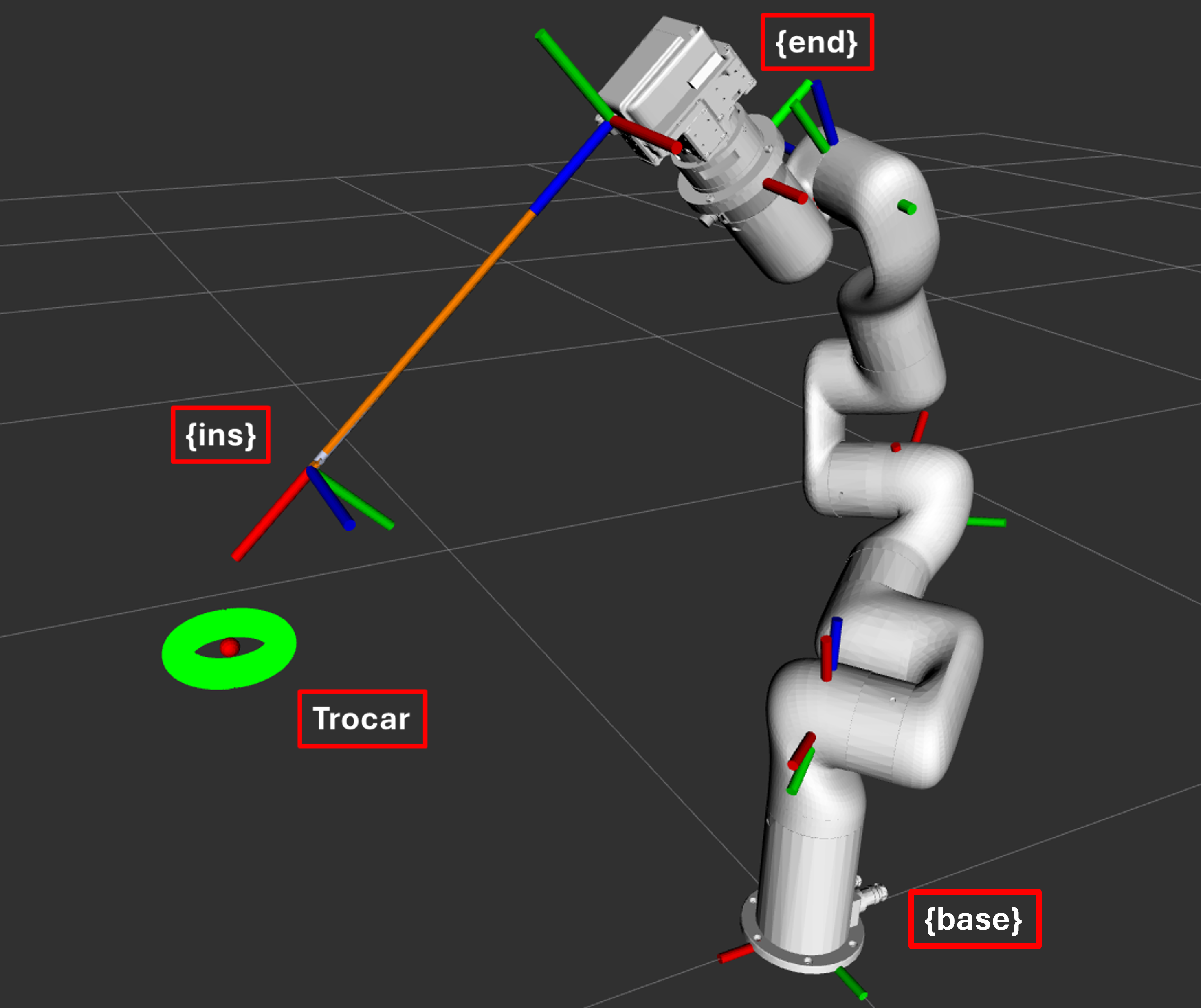}
    \caption{Coordinate frames utilized in the kinematics modeling}
    \label{fig:coordniate_frames}
\end{figure}

\section{Kinematic Modeling} \label{Kinematic}
The robotic system in this work comprises a robot manipulator equipped with an instrument module that drives the dVRK wristed surgical instrument. 
We refer to the robot manipulator's operational (\emph{end}) as the end-effector, and to the central point of the instrument's wrist as the instrument (\textit{ins}), as shown in Fig.~\ref{fig:coordniate_frames}. 
The F/T sensor is positioned at the base of the IM and installed subsequent to the end-effector. For clarity, we have adopted the nomenclature presented in Table \ref{tab:nomeclauture}.

\subsection{Forward Kinematics} \label{dir_kin}

In this section, the direct kinematics and the instantaneous kinematics are derived. The system includes a 7-DoF manipulator and a 4-DoF surgical instrument. Figure~\ref{fig:coordniate_frames} illustrates the coordinate frames defined.

\begin{figure}[!h]
	\centering
	\includegraphics[width=0.49\textwidth, height=0.24\textheight]{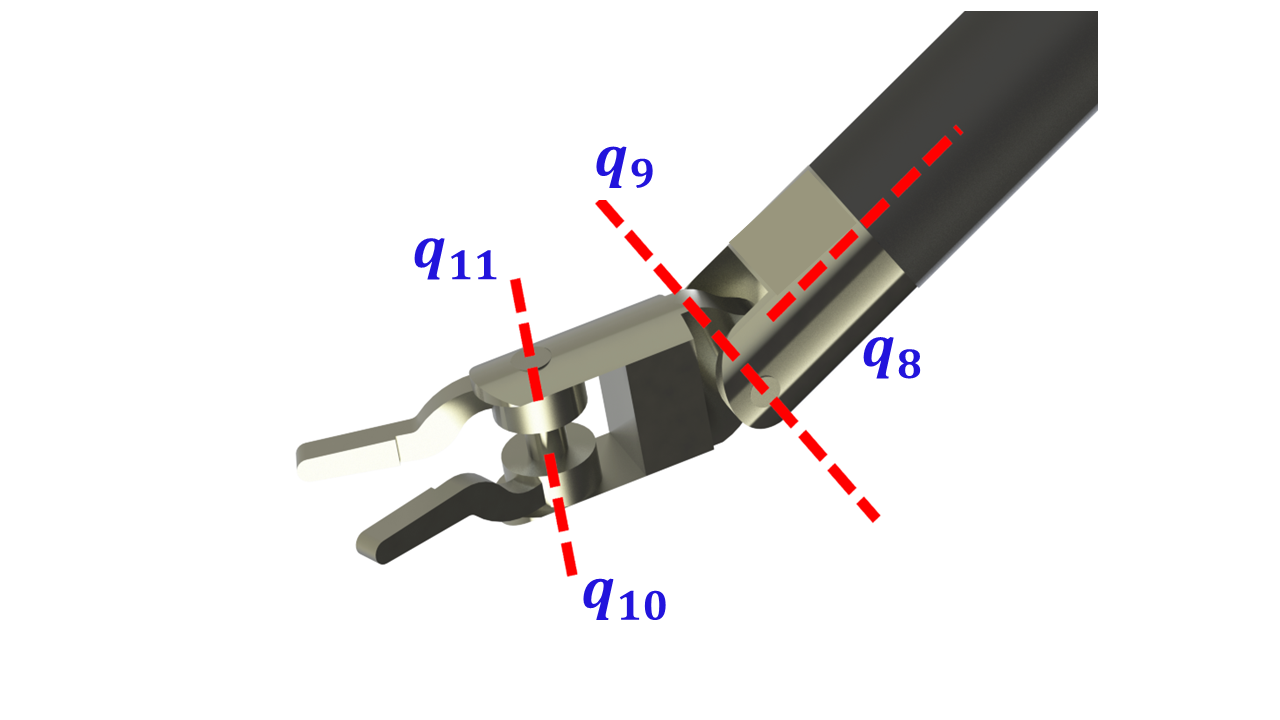}
	\caption{Surgical instrument joints and their axis of rotation}
	\label{fig:ins_joints}
\end{figure}

The current instrument pose is denoted as:
\begin{equation}
\mathbf{x}_{\text{ins,c}} \triangleq (\mathbf{p}_{\text{ins,c}}, \mathbf{R}_{\text{ins,c}})
\end{equation}
where \( \mb{p}_{\text{ins,c}} \) represents the current position of the instrument (ins) and \( \mb{R}_{\text{ins,c}}\) corresponds to the current orientation of the instrument (ins). Or, using a transformation matrix, we have:
\begin{equation}
     ^\text{base}\mathbf{T}_{11} \triangleq \left[\begin{array}{c}
		\mathbf{R}_{\text{ins,c}} \quad \mathbf{p}_{\text{ins,c}} \\
		 \mb{0}_{1\times3} \quad 1
	\end{array}\right]
\end{equation}
The joint variables of the system are defined using a joint space variable vector $\mb{q}$, as:
\begin{gather}
    \mb{q} \triangleq \left[\;\mb{q}_\text{arm}^T, \;\mb{q}_\text{ins}^T\;\right] ^T \\
    \mb{q}_\text{arm} = [q_1,q_2,..., \hspace{1pt}q_7]^T, \quad 
    \mb{q}_\text{ins} = [q_{8},...,\hspace{1pt}q_{11}]^T
\end{gather}
where the joint subspace vectors, $\mb{q}_\text{arm}$ and $\mb{q}_\text{ins}$, respectively, describe the 7-DoF robot manipulator joints and the 4-DoF surgical instrument joints. The instrument joints are illustrated in Fig.~\ref{fig:ins_joints}, and these joints are actuated by four servo motors. 
Then, the position and orientation of the instrument are given by:
\begin{equation}
    ^\text{base}\mathbf{T}_{11} = \prod_{k=1}^{11} \; {}^{k-1}\mathbf{T}_{k} (q_i)
\end{equation}
where \(^{k-1}\mathbf{T}_{k}\) indicates the transformation between frames \(\{k-1\}\) and \(\{k\}\), and it is a function of $q_i$. \par 
The instantaneous kinematics is formulated using a Jacobian matrix as below to relate the joint velocities to the instrument twist \(\boldsymbol{\xi}_{\text{ins}}\):
\begin{align}
& \boldsymbol{\xi}_{\text{ins}} \triangleq \mathbf{J}_{\text{ins}} \; \dot{\mathbf{q}} \hspace{2pt},  \hspace{12pt} \quad \boldsymbol{\xi}_{\text{ins}} \in \mathbb{R}^{6} \label{eq:ins_twist_jacob} \\
& \mathbf{J}_{\text{ins}} = \begin{bmatrix}
\mathbf{J}_{p} \\
\mathbf{J}_{o}
\end{bmatrix}\\
& \boldsymbol{\xi}_\text{ins} \triangleq \begin{bmatrix}
{\mb{v}}^T  , {\boldsymbol{\omega}}^T
\end{bmatrix}^T \label{eq:ins_twist}
\end{align}
The linear and angular parts, $\mb{J}_p$ and $\mb{J}_o$ respectively, are given by:
\begin{align}
& \mathbf{J}_{p} = 
\Big[
\begin{array}{c;{2pt/2pt}c;{2pt/2pt}c}
\mb{J}_{p_1} & \cdots & \mb{J}_{p_{11}}
\end{array}
\Big] \\[4pt]
& \mathbf{J}_{o} = 
\Big[
\begin{array}{c;{2pt/2pt}c;{2pt/2pt}c}
\mb{J}_{o_1} & \cdots & \mb{J}_{o_{11}}
\end{array}
\Big] \\[4pt]
& \mathbf{J}_{p_k} = \hat{\mb{o}}_{k} \times (\mathbf{p}_{\text{ins,c}} - \mathbf{p}_{k}), \quad
\mathbf{J}_{o_k} =\hat{\mb{o}}_{k},
\end{align}
where \( \mb{p}_{k} \) denotes the position of frame \{k\}, and \( \hat{\mb{o}}_{k} \) indicates the axis of rotation of the corresponding joint.
\par
\subsection{Constrained Kinematics} \label{RCM_kin}
This paper builds on our prior work in \cite{nasiri2024}.
The position of the RCM is prioritized to maintain the stability and precision of the entry point during RAMIS. The orientation of the instrument shaft, on the other hand, can be adjusted dynamically to suit the surgical task. We express the position of the RCM using  the end-effector position \(\mb{p}_{\text{end}}\) and the instrument position \(\mb{p}_{\text{ins}}\):
\begin{equation} 
	\mb{p}_{\text{rcm}} = \mb{p}_\text{end} + \lambda \,(\mb{p}_\text{ins} - \mb{p}_\text{end}), \quad \lambda\in (0,1)
	\label{eq:def_RCM}
\end{equation}
Taking derivative, we derive the RCM Jacobian:
\begin{equation}
\mb{\dot{p}}_{\text{rcm}} =
\underbrace{
\Big[
\begin{array}{c;{2pt/2pt}c}
(\mb{J}_\text{end} + \lambda \,(\mb{J}_\text{ins} - \mb{J}_\text{end})) & \mb{d}_\text{ins}
\end{array}
\Big]
}_{\triangleq\; \mb{J}_{\text{rcm}}}
\underbrace{
\begin{bmatrix} \mb{\dot{q}} \\ \dot{\lambda} \end{bmatrix}
}_{\dot{\mb{q}}_{\text{aug}}}
\label{RCM_velocity}
\end{equation}
Here, $\mb{d}_\text{ins}$ represents the relative position of the instrument with respect to the end-effector (a vector that points from $\mb{{p}}_{\text{end}}$ to $\mb{{p}}_{\text{ins}}$).
\begin{equation}
\mb{d}_\text{ins} = \mb{p}_\text{ins} - \mb{p}_\text{end}   \label{eq:d_ins}
\end{equation}
%
%
$\mb{J}_{\text{ins}}$ and $\mb{J}_{\text{end}}$ are the manipulator Jacobians mapping joint velocities to the velocities of the robot's end-effector and instrument as,
\begin{align}
\mb{\dot{p}}_{\text{end}} &= \mb{J}_{\text{end}} \;\mb{\dot{q}} \label{end_velocity} \\
\boldsymbol{\xi}_{\text{ins}} &= \mb{J}_{\text{ins}} \;\mb{\dot{q}} \label{ins_velocity}
\end{align}\par
For derivation convenience, we define an \emph{augmented} joint space vector, $\mb{q}_\text{aug}$,
\begin{equation}
    \mb{q}_\text{aug} \triangleq \left[\mb{q}^T,\;\dot{\lambda}\right]^T
\end{equation}
We define the RCM constraint and the commanded motion constraint for the instrument using the following velocity constraints:
\begin{align}
	s.t.\quad
	& \mb{J}_\text{rcm}
	\begin{bmatrix}
		\dot{\mb{q}}\\
		\dot{\lambda}
	\end{bmatrix} \; = \dot{\mb{p}}_\text{rcm} \label{eqn:RCM_constraint}\\[2pt]
	s.t.\quad & \mb{J}_\text{ins}
	\;\dot{\mb{q}}
	\;= {\boldsymbol{\xi}}_{\text{ins}}\label{eqn:ins_constraint}
\end{align}

Integrating both constraints through an augmented Jacobian formulation,
\begin{equation}
	s.t.\quad
	\mb{J}_\text{total}
	\begin{bmatrix}
		\dot{\mb{q}}\\
		\dot{\lambda}
	\end{bmatrix} \; = \boldsymbol{\xi}_{\text{aug}} \label{eqn:total_constraint} 
\end{equation}
where,
\begin{align}
	& \mb{J}_{\text{{total}}} = \left[\begin{array}{c}
		\mb{J}_{\text{{ins}}} \quad \mb{0}_{6\times1} \\
		\hdashline
		\mb{J}_{\text{{rcm}}}
	\end{array}\right], \quad \mb{J}_{\text{{total}}}\in\mathbb{R}^{9\times12}
	\label{eq:total_jacobian} \\[2pt]
	&     
	\boldsymbol{\xi}_{\text{aug}} = 
	\left[\begin{array}{c}
		\boldsymbol{\xi}_\text{ins}\\
		\hdashline
		\dot{\mb{p}}_{\text{rcm}} 
	\end{array}\right],  \qquad \boldsymbol{{{\xi}}}_{\text{aug}} \in \mathbb{R}^{9\times 1}
	\label{eqn:x_cmd_velocity}
\end{align}

\subsection{Redundancy Resolution} \label{redundacny_res}
A general solution satisfying the constraint in \eqref{eqn:total_constraint} can be expressed as:
\begin{equation}
        \underbrace{\begin{bmatrix}
            \dot{\mathbf{q}} \\
            \dot{\lambda} 
        \end{bmatrix}_{\text{}}}_{\mathbf{\dot{\mb{q}}_\text{aug}}} = \mathbf{J}^{\dagger}_{\text{total}} \hspace{2pt} \boldsymbol{\xi}_{\text{aug}} + \left(\mathbf{I}-\mathbf{J}^{\dagger}_{\text{total}} \; \;\mathbf{J}_{\text{total}}\right) \boldsymbol{\eta}, \quad 
        \bs{\eta}\in\mathbb{R}^{12} 
    \label{redundancy_without_admt} 
\end{equation} \par
The system has a redundancy of 2-DoF, and it can be explained as follows. The manipulator has 7-DoF. While the instrument is a 4-DoF mechanism, the gripper open/close DoF does not contribute to task space, which then makes the the instrument equivalent to a 3-DoF one. The RCM requires a 2-DoF constraint while the user command specifies a 6-DoF motion constraint. \par
This redundancy can be resolved by selecting a vector \(\boldsymbol{\eta}\) in \eqref{redundancy_without_admt}, which represents an arbitrary choice that can fulfill the task constraint. Vector $\bs{\eta}$ can be designed as the gradient of a cost function of interest. \\
%
The upper part of the augmented twist in \eqref{eqn:x_cmd_velocity} represents a \(6 \times 1\) velocity vector of the instrument.
 This vector is derived from the resolved motion rate algorithm and obtained via the haptic device user interface command. The lower part of \eqref{eqn:x_cmd_velocity}, on the other hand, represents the RCM velocity and can be derived from a proposed admittance controller mention in \eqref{RCM_velocity2}.
The final augmented joint velocities command \eqref{redundancy_without_admt} can be rewritten as:
\begin{equation}
    \begin{split}
        \mathbf{\dot{\mb{q}}_\text{aug}} &= \mathbf{J}_{\text{total}}^{\dagger} \hspace{2pt} \boldsymbol{{{\xi}}}_{\text{aug}} + \left(\mathbf{I}-\mathbf{J}_{\text{total}}^{\dagger} \mathbf{J}_{\text{total}}\right) \boldsymbol{\eta}
    \end{split}
    \label{eq:final_vel_command} 
\end{equation}
where \(\mathbf{J}_{\text{total}}^{\dagger}\) represents the Moore-Penrose pseudo-inverse of the Jacobian, \(\mathbf{J}_{\text{total}}\), as shown in~\eqref{eq:total_jacobian}. 
The vector \(\boldsymbol{\eta}\) mentioned in \eqref{eq:final_vel_command} represents a null-space  cost gradient of interest. We chose the interpolation variable $\lambda$ to be as close as possible to a fixed and arbitrary reference value $\lambda_0$, for experimental validation. The residue vector \(\boldsymbol{\eta}\) is then defined as follows:
\begin{align}
&\boldsymbol{\eta} = \nabla_{(\mb{q}, \mb{\lambda})} \left( \frac{1}{2} \left\| \mb{\lambda} - \mb{\lambda}_0 \right\|^2 \right) \bigg|_{\mb{\lambda}_0 = 0.4}, \\
&\boldsymbol{\eta} = \; \; \big[\mb{0}_{1\times11}, \;\lambda - 0.4\big]^T
\end{align}
Finally, the joint commands for the manipulator can then be found from \eqref{eq:final_vel_command},
\begin{equation}
    \mb{q}= \mathbf{\dot{\mb{q}}_\text{aug}} \hspace{2pt} \text{dt} +\mb{q} 
\end{equation}

\begin{figure*}[!h]
    \centering
    \includegraphics[width=\linewidth]{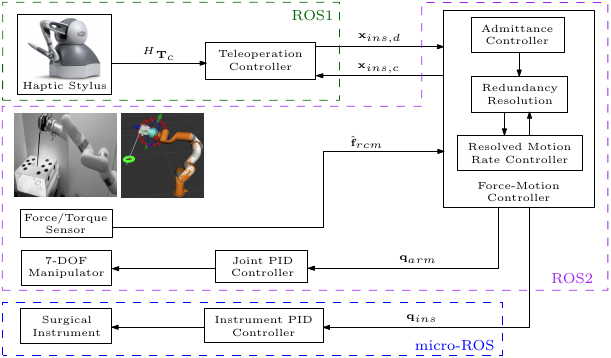}
    \caption{Teleoperation control architecture of the surgical manipulator system.}
    \label{fig:control_schematic}
\end{figure*}

\subsection{Admittance Control} \label{admitance_controller}
We introduced an augmented command velocity constraint $\bs{\xi}_\text{aug}$ in \eqref{eqn:x_cmd_velocity}, which contains $\dot{\mb{p}}_\text{rcm}$. In this section, we integrate an admittance controller that specifies the RCM point velocity $\dot{\mb{p}}_\text{rcm}$ based on the estimated force $\hat{\mb{f}}_\text{rcm}$. An admittance law is proposed in \eqref{RCM_velocity2}.
\begin{equation}
     \mb{\dot{p}}_{\text{rcm}} = K_\text{adm} \; \left(\mb{I} - \bs{\Omega}\right) \; \;{\mb{f}}_{\text{e,rcm}}\label{RCM_velocity2}
\end{equation}
where,
\begin{align} 
    & \bs{\Omega} = {\mb{n}_d}\; \left({\mb{n}_d}^T\right), \qquad \mb{n}_d = \frac{\mb{d}_\text{ins}}{\|\mb{d}_\text{ins}\|} \label{eqn:Omega_proj}\\[2pt]
    & {\mb{f}}_{\text{e,rcm}}=\hat{\mathbf{f}}_{\text{rcm}} - {\mathbf{f}}_{\text{rcm,desired}}\label{force_error}
\end{align}
In \eqref{eqn:Omega_proj}, a projection matrix, $\bs{\Omega}$, is constructed. Thereby, in \eqref{RCM_velocity2}, a null space projector $\left(\mb{I} - \bs{\Omega}\right)$ can be used to calculate a force projected onto a plane that is perpendicular to the instrument shaft and that is located at the RCM.  A scalar admittance gain, $K_\text{adm}$, is used. The vector, ${\mb{f}}_{\text{e,rcm}}$, represents the force tracking error at the RCM. 
For derivation convenience, a projected admittance gain matrix is denoted as follows:
\begin{equation}
    \mb{K}_{\text{adm}\perp} = K_\text{adm} \; \left(\mb{I} - \bs{\Omega}\right), \quad \mb{K}_{\text{adm}\perp} \in \mathbb{R}^{3\times3}
\end{equation} \par

\section{Teleoperation} \label{haptic}
In this study, we present a leader-follower teleoperation framework designed for direct user-supervised motion control of surgical instruments. The framework, depicted in Fig.~\ref{fig:control_schematic}, involves user interaction with a haptic device, transmitting the haptic stylus's pose to the teleportation controller. The teleoperation controller also acquires the current instrument pose from the main force-motion controller in \emph{ROS2}. Subsequently, it computes a relative haptic device command pose from an anchor pose to its current position and sends this command pose as the desired instrument pose to the leader robot system.

\subsection{Haptic Device as User Interface}
\begin{figure}[!hb]
	\centering
	\includegraphics[width=0.49\textwidth, height=0.27\textheight]{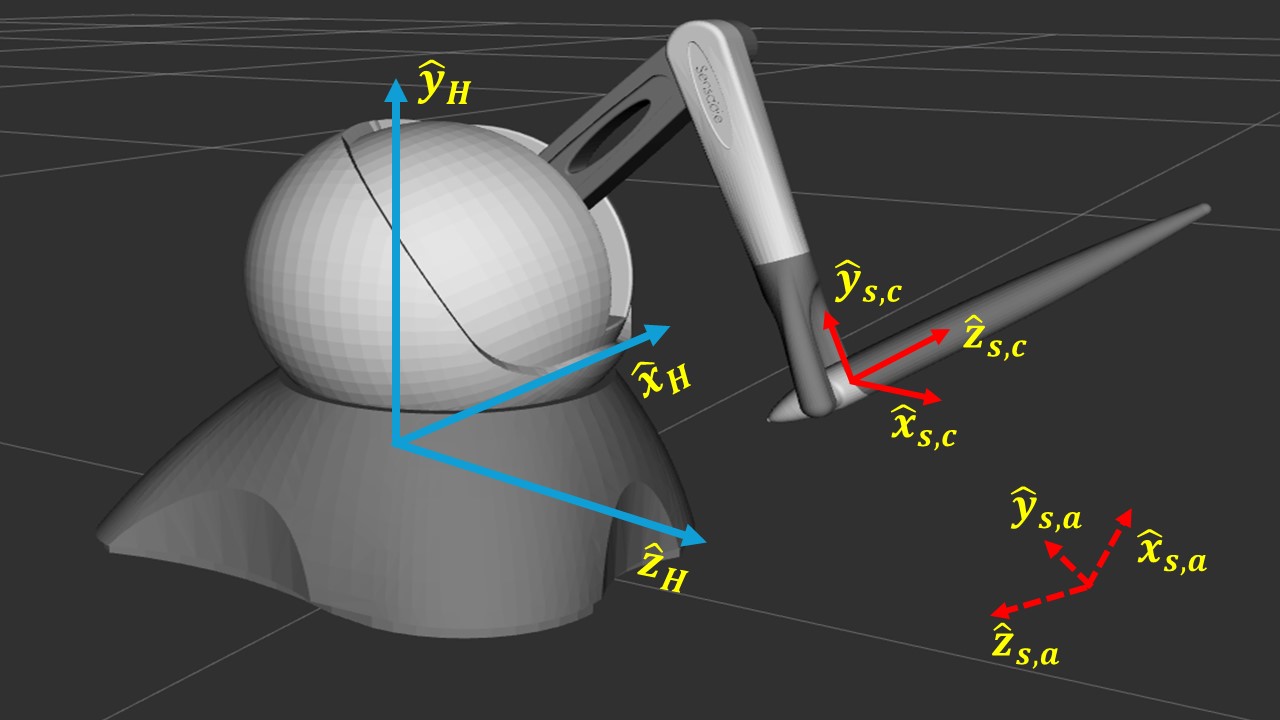}
	\caption{3D Systems Touch haptic stylus reference frames.}
	\label{fig:haptic_frames}
\end{figure}
In this paper, a haptic stylus is employed as the leader device for teleoperating the surgical manipulator system. The stylus supports 6 degrees of freedom (DoF), enabling intuitive control of the follower manipulator. Due to its widespread adoption in haptics and teleoperation research, open-source APIs facilitate communication with the haptic stylus for researchers. The software package discussed in \cite{CRTK2020} includes a \emph{ROS} driver that transmits pose messages from the haptic stylus, along with the state of its two buttons, at 500Hz.

\subsection{Trajectory Mapping}
The proximal button on the stylus is programmed for use as the teleoperation command button for the surgical manipulator. The pose of the stylus when teleoperation is initiated is called the anchor pose, registered as \(^{H}\mathbf{T}_{s,a}\). All subsequent poses when the command button is engaged is referred to as the current pose, denoted by \(^{H}\mathbf{T}_{s,c}\). An example visual representation of these frames can be seen in Fig. \ref{fig:haptic_frames}. The relationship between the anchor and current pose of the stylus is as follows:
\begin{equation}
    ^{H}\mathbf{T}_{s,c} = {}^{H}\mathbf{T}_{s,a}\;{}^{s,a}\mathbf{T}_{s,c/a}\;,
\end{equation}
where \(^{s,a}\mathbf{T}_{s,c/a}\) represents the change in pose from the anchor frame to the current frame of the stylus, as represented in the anchor frame. During teleoperation, we are interested in retrieving this relative change in pose as it corresponds to the surgical manipulator's relative change in pose through a similarity transformation.
\begin{equation}
    ^{\text{ins}, a}\mathbf{T}_{s,c/a} = ({}^{\text{ins}, a}\mathbf{T}_{s,a})({}^{s,a}\mathbf{T}_{s,c/a}) ({}^{\text{ins}, a}\mathbf{T}_{s,a})^{-1}\;,
\end{equation}

\noindent where \(({\text{ins},a})\) is the instrument \emph{anchor} frame, which corresponds to the surgical manipulator's pose when the command button is initially pressed. \({}^{\text{ins}, a}\mathbf{T}_{s, a}\) is computed using the following expression:
\begin{equation}
    ^{\text{ins}, a}\mathbf{T}_{s,a} = ({}^\text{base}\mathbf{T}_{\text{ins}, a})^{-1} ({}^\text{base}\mathbf{T}_{H}) ({}^{H}\mathbf{T}_{s,a})\;
\end{equation}

\noindent where \(^\text{base}\mathbf{T}_{\text{ins},a}\) is the surgical instrument anchor pose as written in the manipulator's base frame.

This transformed relative pose is essential to compute the pose of the surgical manipulator written in its base frame using the following expression:
\begin{equation}
    ^\text{base}\mathbf{T}_{\text{ins},d} = {}^\text{base}\mathbf{T}_{\text{ins}, a}\;{}^{\text{ins}, a}\mathbf{T}_{s,c/a}\; \label{eq:teleop-desired}
\end{equation}

\noindent where \(^\text{base}\mathbf{T}_{\text{ins},d}\) is a homogeneous transformation matrix representing the desired pose of the instrument written in the surgical manipulator base frame.

\subsection{Resolved Rate Algorithm} \label{RR_algthm}
Consider the desired pose of the robot instrument defined as follows:
\begin{equation}
\mathbf{x}_{\text{ins,d}} \triangleq (\mathbf{p}_{\text{ins,d}}, \mathbf{R}_{\text{ins,d}})
\end{equation}
This is specified in real-time by the operator and managed through a resolved motion-rate algorithm that processes joint commands for controlling the manipulator, initially introduced in \cite{Whitney1969}.
The position error \( \mb{e}_{p} \) of the robot instrument is defined as the difference between the current pose \( \mb{x}_{\text{ins,c}} \) and the desired pose \( \mb{x}_{\text{ins,d}} \):
\begin{align}
\mb{e}_p &= \mb{p}_{\text{ins,d}} - \mb{p}_{\text{ins,c}}
\end{align}
The orientation error \( \mb{e}_{\mu} \) of the manipulator instrument is computed by first calculating the relative error rotation matrix \( \mb{R}_{e} \), which represents the difference between the current orientation \( \mb{R}_{\text{ins,c}} \) and the desired orientation \( \mb{R}_{{\text{ins,d}}} \):

\begin{align}
& \mb{R}_e = \mb{R}_{\text{ins,d}} \hspace{1pt} {\mb{R}^T_{\text{ins,c}} } \\
& \theta = \arccos\left(\frac{\text{tr}(\mb{R}_e) - 1}{2}\right) \\
& \mb{e}_\mu = \frac{\theta}{2 \sin \theta} 
\begin{bmatrix}
    \mb{R}_e(3,2) - \mb{R}_e(2,3) \\
    \mb{R}_e(1,3) - \mb{R}_e(3,1) \\
    \mb{R}_e(2,1) - \mb{R}_e(1,2)
\end{bmatrix}
\end{align}
\begin{align}
v_{\text{m}} &= 
\begin{cases} 
    V_{\text{max}} & \text{if } \|\mb{e}_p\| > \frac{\epsilon_p}{\gamma_p} \\
    V_{\text{min}} + ( V_{\text{max}} - V_{\text{min}}) \beta_v & \text{if } \|\mb{e}_p\| \leq \frac{\epsilon_p}{\gamma_p}
\end{cases} \\
\omega_{\text{m}} &= 
\begin{cases} 
    \omega_{\text{max}} & \text{if } \|\mb{e}_\mu\| > \frac{\epsilon_\mu}{\gamma_\mu} \\
    \omega_{\text{min}} + (\omega_{\text{max}} - \omega_{\text{min}}) \beta_\omega & \text{if } \|\mb{e}_\mu\| \leq \frac{\epsilon_\mu}{\gamma_\mu}
\end{cases}
\end{align}
where, 
\begin{equation}
    \beta_v = \frac{\|\mb{e}_p\| - \gamma_p}{\gamma_p(\epsilon_p - 1)}, \qquad
    \beta_\omega = \frac{\|\mb{e}_\mu\| - \gamma_\mu}{\gamma_\mu(\epsilon_\mu - 1)}
\end{equation}
Thereby, the desired linear and angular velocities are given by:
\begin{align}
&\mb{v} = \frac{v_{\text{m}} \, \mb{e}_p}{\|\mb{e}_p\|} \label{eq:v_twist}\\
&\boldsymbol{\omega} = \frac{\omega_{\text{m}} \, \mb{e}_\mu}{\|\mb{e}_\mu\|} \label{eq:w_twist}
\end{align}

Resolved motion-rate control uses position and orientation errors to compute the desired instrument twist, \( \boldsymbol{\xi}_\text{ins} \), as shown in \eqref{eq:ins_twist}. As the \textit{instrument} gets closer to the target position and the errors converge to zero, the instrument twist reduces.\\
Here, \(V_{\text{max}}\) and \(V_{\text{min}}\) denote the maximum and minimum linear velocities of the instrument, while \(\omega_{\text{max}}\) and \(\omega_{\text{min}}\) indicate the maximum and minimum angular velocities, respectively.
\(\gamma_p\) and \(\gamma_\mu\) represent the allowable thresholds for position and orientation errors, while \(\epsilon_p\) and \(\epsilon_\mu\) signify the relative error radii within which the instrument transitions from a resolved rate to a reduced proportional rate.\par

\subsection{System Control Architecture}

Figure \ref{fig:control_schematic} shows the control architecture for teleoperation of the surgical manipulator system. The haptic stylus sends the device state information to the teleoperation controller that works within a ROS1 environment. The force-motion controller encompasses the admittance control and the resolved motion rate control employing a redundancy resolution method, all within a ROS2 environment to communicate with the 7-DoF manipulator and the F/T sensor. Communications between the teleoperation controller and the force-motion controller are facilitated by a ROS1-ROS2 bridge. The instrument module (IM) is actuated by servos controlled using micro-ROS, extending ROS2 functionalities to the microcontroller. The control system has been designed for precise and responsive teleoperational control, and its modular architecture integrating ROS1 and ROS2 enables us to leverage existing open-source software packages for real-time control.

\section{Instrument Module} \label{IDM}
\begin{figure*}[!t]
    \centering
    \renewcommand{\thesubfigure}{\alph{subfigure}}
    \captionsetup[subfigure]{labelformat=simple, labelsep=none}
    \begin{subfigure}[b]{2\columnwidth}
        \centering
        \includegraphics[width=\linewidth, height=0.31\textheight]{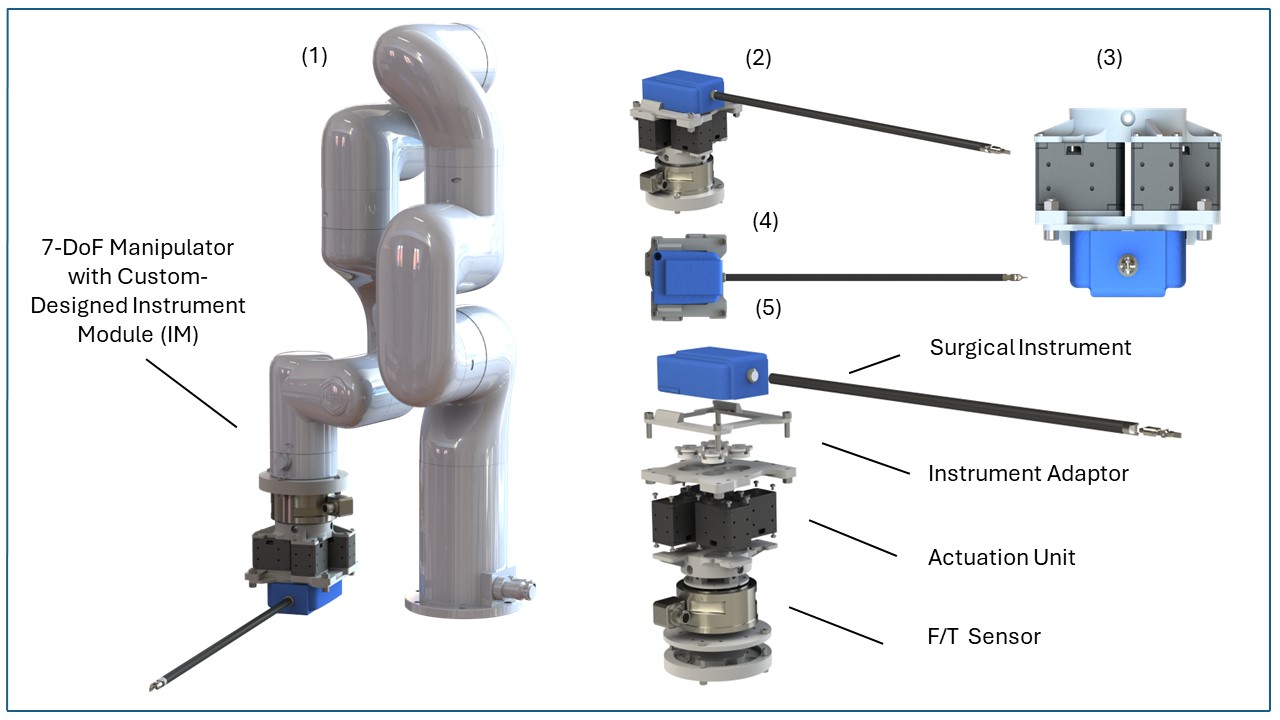}
        \caption{) A 4-DoF actuation unit with the attached F/T sensor for cable-driven surgical instrument: (1) the assembled CAD model of the 7-DoF Manipulator with the custom-designed IM; (2) the assembled CAD model of the IM; (3),(4) Top and front views of the module; and (5) the exploded view.}
        \label{fig:Multiple_views}
    \end{subfigure}
    \vfill
    \begin{subfigure}[b]{2\columnwidth}
        \centering
        \includegraphics[width=\linewidth, height=0.3\textheight]{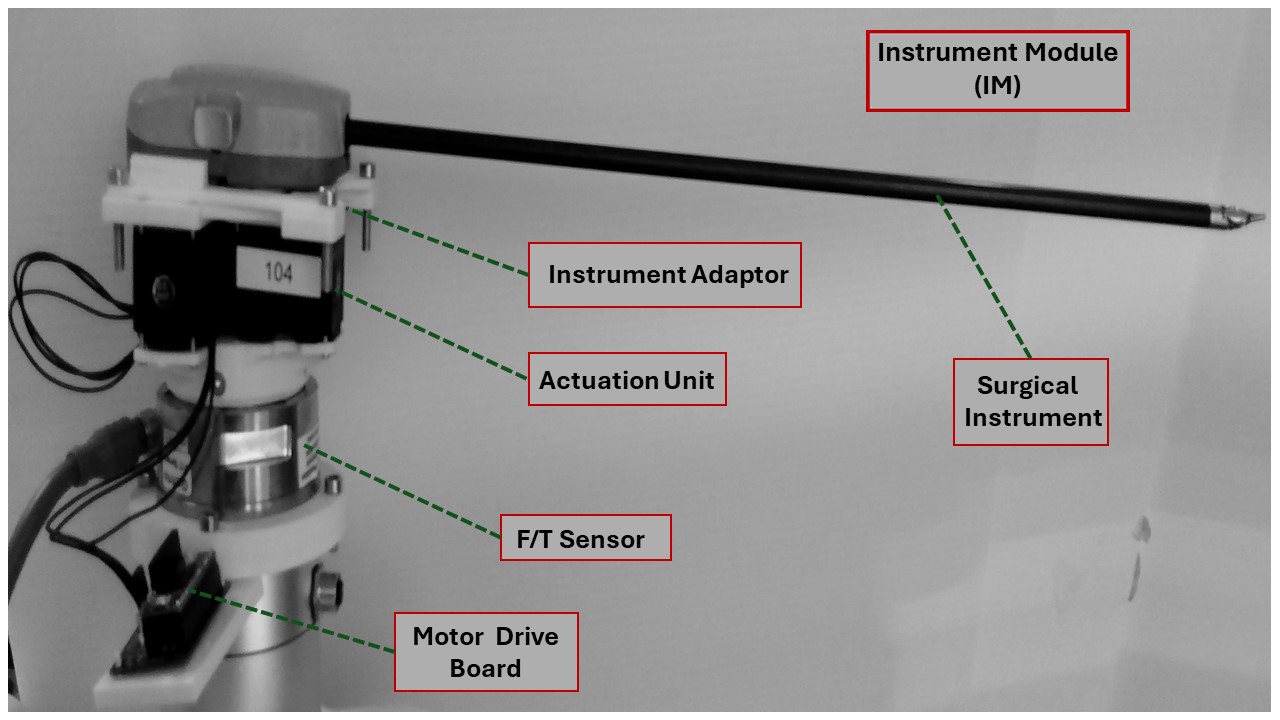}
        \caption{) IM Architecture Design Including: off-the-shelf surgical instrument, actuation unit, instrument adapter, F/T sensor and motor driver board.}
        \label{fig:IDM_arch}
    \end{subfigure}
    \caption{CAD and 3D-printed models of the proposed Instrument Module (IM)}
    \label{fig:combined}
\end{figure*}

The function of the \emph{Instrument Module} (IM) is to facilitate articulation control of the surgical instrument, as presented in Fig.~\ref{fig:combined}. The mechatronics development of the IM are explained in this section.
\subsection{Design Features}

The proposed low-cost IM for driving a cable-driven dVRK instrument includes the following design features:
\begin{itemize}
     \item \textit{Compactness}: The module is lightweight and compact, making it suitable for mounting on and manipulation by a collaborative robot.
     
    \item \textit{Mechanical Compatibility}: The module effectively actuates various types of cable-driven dVRK surgical instruments (up to 4-DoF), while allowing for a simple, adaptable mechanical design. This ensures consistent performance and reduces development time.
    
    \item \textit{ROS Compatibility}: The module is developed within the micro-ROS infrastructure, reducing the delay between operator commands and the real-time actuation of the surgical instrument during teleoperation.
\end{itemize} \par

\subsection{Mechanical Design}
The actuation unit drives the surgical instrument via a four-pulley interface.
The proposed module has four DoFs actuation functionality provided by four Dynamixel XL-430 servo motors. Figure~\ref{fig:Multiple_views} shows the exploded view of the CAD assembly with an F/T sensor attached at the base of the IM and the surgical instrument. The design specifications of the proposed IM are listed in Table~\ref{tab:param}. The hardware prototype of the IM is presented in Fig.~\ref{fig:Multiple_views}, and the CAD files of the design are open sourced\footnote{\url{https://github.com/stevens-armlab/Teleoperation_ISI_IM}}.
\begin{table}[ht]
	\centering
   \captionsetup{justification=centering}
	\caption{Instrument module specifications}
	\begin{tabular}{ p{0.45\columnwidth}|p{0.45\columnwidth} }
		\hline
		\textbf{Component} &  \textbf{Specification} \\
		\hline
		Servo model &  Dynamixel XL-430-W250 \\[1.5pt]
		Motor adaptor Dimension &  89.7 $\times$ 85.6 (mm) \\[1.5pt]
		Diameter of the motor pulley & 10.8 (mm) \\[1.5pt]
        Surgical instrument & (dVRK) wristed surgical instrument \\[1.5pt]
		F/T sensor & ATI-Gamma-SI-130-10 \\
		\hline
	\end{tabular}
	\label{tab:param}
\end{table}
\subsection{Electronics Development}
As detailed in Section \ref{haptic} and illustrated in Fig.~\ref{fig:control_schematic}, the haptic device facilitates user commands for the instrument that is held by the robot manipulator. The main control node and resolved rate algorithm run in ROS2, while the haptic device uses ROS1 to send the desired instrument pose to the ROS2 node via a ROS1-ROS2 bridge. A Teensy board and Micro-ROS facilitate communication with the ROS2 node for the instrument's actuation unit, controlling the servos.
Micro-ROS adapts the robot operating system framework for microcontrollers. This integration allows the Teensy board to communicate with ROS2, enabling servo control and, consequently, the control of the surgical instrument.

We used the \textit{Teensy}  board as the driving board for the actuation unit. A custom Dynamixel shield was utilized with the Teensy board, employing UART serial communication for reliable and high-speed control of the servos. Additionally, we set up \textit{Teensyduino}, an add-on for the Arduino software, to program and use the Teensy series of microcontrollers within the Arduino IDE.

\section{Experimental Validation} \label{validation}
The telemanipulation system depicted in Fig.~\ref{fig:System_arch} was evaluated for its proficiency in performing a few surgical tasks.
We tested the system under various conditions and tasks, within the typical allowable four degrees of freedom in laparoscopic surgery: movement along the insertion axis and three tilt angles (roll, pitch, yaw). As shown in Fig.~\ref{fig:System_arch}, we tested the system using a surgical simulator kit to assess its capability to accomplish the following tasks: (1) pick and place, (2) passing a thread through a ring, and (3) trajectory tracking. 
\subsection{Pick and Place and Thread Passing} \label{pck_n_plce}
\begin{figure}
	\centering
	\includegraphics[width=0.48\textwidth, height=0.29\textheight]{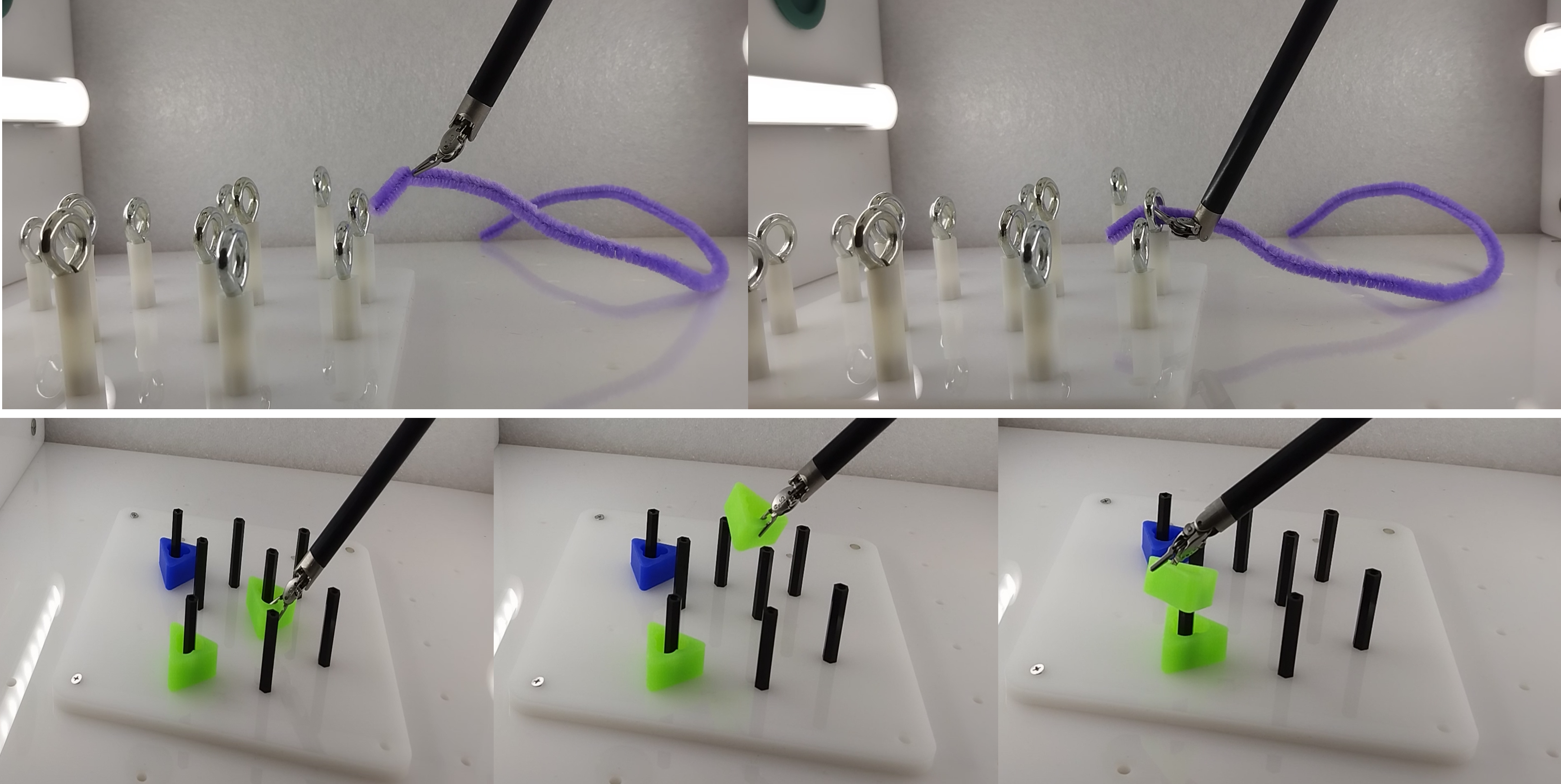}
	\caption{\text{Top}: Passing a thread through the ring, \text{Bottom}: Pick-and-place actions utilizing the pegboard.}
	\label{fig:pcknplace}
\end{figure}

Evaluations for pick-and-place task involved handling plastic triangular-shaped objects using a pegboard. The thread-passing task involved passing a thread through rings on a ring board, as illustrated in Fig.~\ref{fig:pcknplace}. \par
First, we conducted multiple tests with various combinations of objects and pegs on the pegboard. The results showed that the tasks were completed successfully and accurately.
The \textit{pegboard} has 8 pegs, each with a diameter of 4.7 mm. The plastic blue and green-colored triangles have a hole with a diameter of 9.4 mm. The experiment involved telemanipulating the triangles by picking them from the pegs at the bottom and placing them on one of the pegs at the top corners.

We also conducted another experiment using a \textit{ring board} which has the same dimensions as the pegboard. The thread board includes 14 rings, each with a diameter of 8.1 mm. The test involved picking up a 3.2mm diameter thread and passing it through multiple rings.

\subsection{Trajectory Tracking} \label{traj_track}
\begin{figure}[!t]
    \centering
    \renewcommand{\thesubfigure}{\alph{subfigure}}
    \captionsetup[subfigure]{labelformat=simple, labelsep=none}
    \begin{subfigure}[b]{1\columnwidth}
        \centering
        \includegraphics[width=\textwidth, height=0.28\textheight]{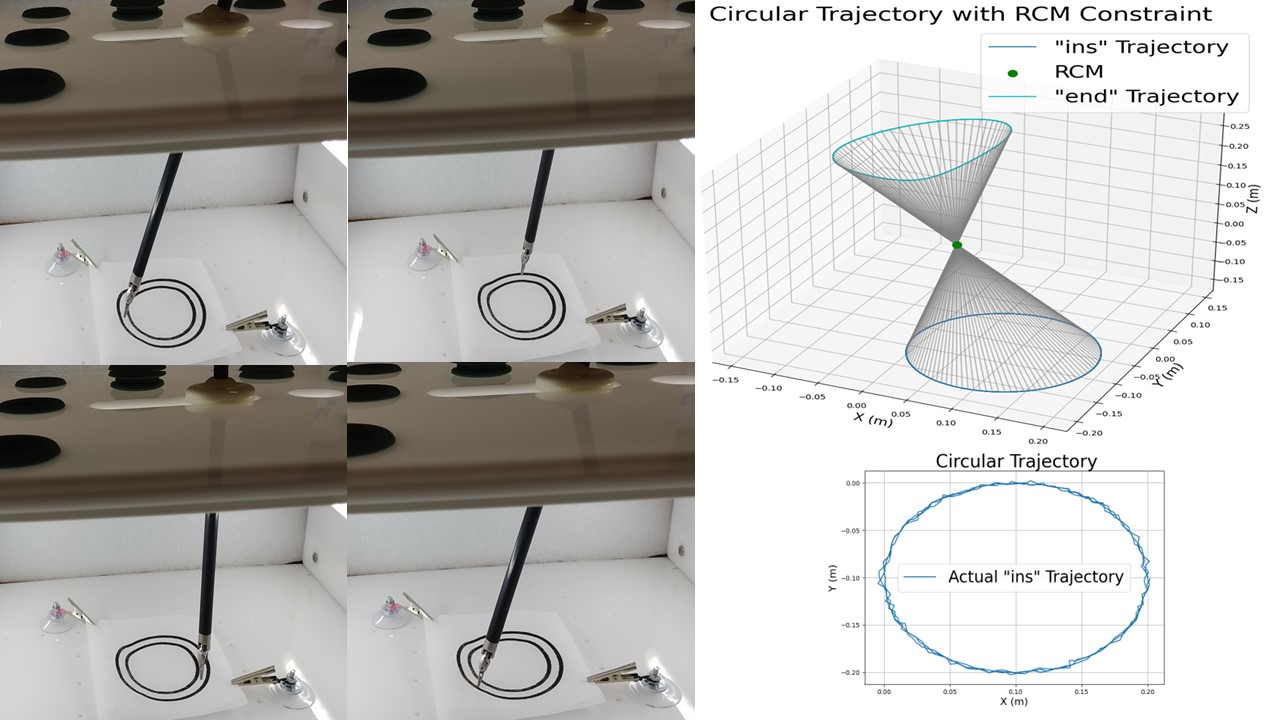}
        \caption{) Sequence of the circular trajectory tracking of the instrument (\textit{ins}), and the actual versus commanded plots.}
        \label{traject_sim_circ}
    \end{subfigure}
    \hfill
    \begin{subfigure}[b]{1\columnwidth} 
        \centering
        \includegraphics[width=\textwidth, height=0.27\textheight]{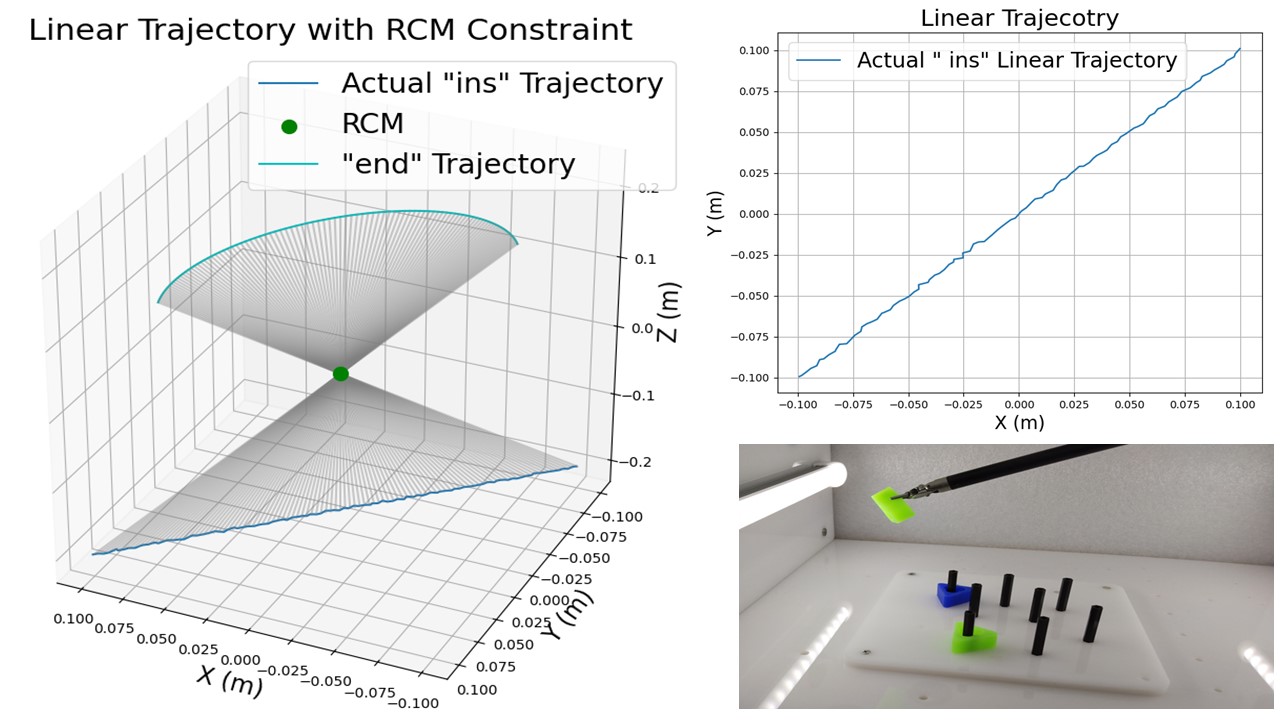}  
        \caption{) Linear trajectory tracking of the robot instrument (\textit{ins}), and the related actual versus commanded plots.}
        \label{traject_sim_lin}
    \end{subfigure}
    \caption{Sequence of the circular and linear teleoperation trajectory tracking of the instrument (\textit{ins}), with the corresponding actual vs. commanded plots.}
    \label{fig:traject_sim}
\end{figure}

We tested our framework's capability to follow the desired teleoperation trajectories. First, we used a circular path with a 10 cm radius for the instrument to follow, and then we plotted the actual path against the haptic device command. As shown in Fig. \ref{traject_sim_circ}, the robot performed the task with a reasonable accuracy. In Fig. \ref{traject_sim_lin}, the robot picks up a triangle from the pegboard and follows a linear path of 20 cm. The plot displays the actual path follows the commanded trajectory with high accuracy. Videos of the demonstrations are made available on YouTube~\footnote{
    \url{https://youtu.be/i2yayaXupwk}, 
    \url{https://youtu.be/yJWDZzUG14I},  
    \url{https://youtu.be/2YILVaBRyWc}
}.

\section{Conclusion}\label{conclusion}
This paper addresses the challenges faced during robot-assisted MIS, specifically those where a robotic system and its tools must operate within confined spaces while adhering to the constraint of the RCM. We leverage our developed framework, integrating admittance control into a redundancy resolution method, specifically designed for the kinematic constraints of RCM. To further expand the system, we incorporate a haptic stylus device to facilitate telemanipulation of a robotic system consisting of a 7-DoF manipulator and a 4-DoF surgical instrument. \par
A compact, low-cost and modular custom-designed instrument module (\text{IM}) was developed to facilitate integration between the manipulator, F/T sensor, surgical instrument, and its actuation unit. The assessment and experimental validations demonstrated the system's competence in handling critical tasks in robot-assisted MIS. Due to the IM's mechatronics design, the proposed framework offers advantages in setup time and repeatability of tasks at different poses.\par
Experimental validations successfully demonstrated the teleoperation framework's capabilities. These include picking and placing items, passing a thread through rings, and tracking various trajectories. All tasks adhered to the RCM constraint using the proposed adaptive admittance control framework.\par
%
%

\section{Acknowledgment}
We would like to thank Yihan Huang for his valuable help in the design and development of the instrument module in this project.\par 
This research was supported in part by NSF Grant CMMI-2138896.


\bibliography{sn-bibliography}

\end{document}